\documentclass[conference]{IEEEtran}
\IEEEoverridecommandlockouts
\usepackage{cite}
\usepackage{amsmath,amssymb,amsfonts}
\usepackage{algorithmic}
\usepackage{graphicx}
\usepackage{textcomp}
\usepackage{xcolor}

\usepackage{pgfplots}
\usepackage{subcaption}
\DeclareMathOperator*{\argmax}{arg\,max}
\usepackage{tikz}
\usetikzlibrary{shapes}
\usepackage{varwidth}
\usepackage{wrapfig}
\usepackage{array,multirow,makecell}

\usepackage{amsmath}
\usepackage{tikz}
\usetikzlibrary{shapes}
\usepackage{varwidth}
\usepackage{wrapfig}
\usepackage{array,multirow,makecell}
\usepackage[switch]{lineno}

\usepackage[ruled,vlined,linesnumbered]{algorithm2e}
\newcommand{\nosemic}{\renewcommand{\@endalgocfline}{\relax}}
\newcommand{\dosemic}{\renewcommand{\@endalgocfline}{\algocf@endline}}
\let\oldnl\nl
\newcommand{\nonl}{\renewcommand{\nl}{\let\nl\oldnl}}

\usepackage{pgfplots}
\usepackage{caption}
\usepackage{subcaption}
\usepackage{xcolor,colortbl}
\definecolor{LightCyan}{rgb}{0.88,1,1}
\definecolor{LightCyan2}{rgb}{0.48,0.69,0.99}
\definecolor{linecolor}{rgb}{0.07,0.33,0.66}
\definecolor{bleu}{rgb}{0.07,0.33,0.76}
\definecolor{bleuvictoria}{rgb}{0.03,0.35,0.61}
\definecolor{bleudepths}{rgb}{0.15,0.39,0.32}
\definecolor{bleumalibu}{rgb}{0.12,0.25,0.96}
\definecolor{rouge}{rgb}{0.83,0.08,0}
\definecolor{racingred}{rgb}{0.74,0.09,0.17}
\definecolor{vert}{rgb}{0.12,0.74,0.13}
\definecolor{gold}{rgb}{1,0.84,0}
\definecolor{aurometalsaurus}{rgb}{0.43, 0.5, 0.5}
\definecolor{cadetgrey}{rgb}{0.57, 0.64, 0.69}
\definecolor{lemon}{rgb}{0.94, 0.992, 0.37}
\definecolor{coffee}{rgb}{0.435, 0.305, 0.215}
\definecolor{liver}{rgb}{0.325, 0.294, 0.309}
\definecolor{rust}{rgb}{0.717, 0.155, 0.055}
\definecolor{darkkhaki}{rgb}{0.717, 0.155, 0.055}
\definecolor{tawny}{rgb}{0.74,0.71,0.42}
\definecolor{darkkhaki}{rgb}{0.817, 0.34, 0}
\definecolor{cocoabrown}{rgb}{0.208, 0.157, 0.118}
\definecolor{lilas}{rgb}{0.71, 0.4, 0.82}

\usepackage{amsthm}
\newtheorem{thm}{Theorem}

\def\BibTeX{{\rm B\kern-.05em{\sc i\kern-.025em b}\kern-.08em
    T\kern-.1667em\lower.7ex\hbox{E}\kern-.125emX}}

\title{Subspace-Focused Search Method for Optimal Coalition Structure Generation\\
}

\author{\IEEEauthorblockN{Redha Taguelmimt}
\IEEEauthorblockA{\textit{LIMOS Laboratory} \\
\textit{Clermont Auvergne University,}\\
\textit{CNRS, LIMOS, F-63000}\\
Clermont-Ferrand, France \\
redha.taguelmimt@gmail.com}
\and
\IEEEauthorblockN{Samir Aknine}
\IEEEauthorblockA{\textit{LIRIS Laboratory} \\
\textit{Univ Lyon, UCBL, CNRS, LIRIS,}\\
\textit{UMR5205, F-69622 Villeurbanne}\\
Lyon, France \\
samir.aknine@univ-lyon1.fr}
\and
\IEEEauthorblockN{Djamila Boukredera}
\IEEEauthorblockA{\textit{Laboratory of Applied Mathematics} \\
\textit{Faculty of Exact Sciences}\\
\textit{University of Bejaia}\\
06000 Bejaia, Algeria \\
djamila.boukredera@univ-bejaia.dz}
\and
\IEEEauthorblockN{Narayan Changder}
\IEEEauthorblockA{\textit{TCG Centres for Research and Education in Science and Technology}\\
Kolkata, India \\
narayan.changder@tcgcrest.org}
}
\begin{document}

\maketitle

\begin{abstract}

Coalition formation is a central approach for multi-agent coordination, which has been an active research topic in AI over the past three decades. It involves partitioning agents into coalitions to maximize the system's performance. In this paper, we provide the first algorithm based on parameter configuration for optimal coalition structure generation. We analyze how parts of the solution space can be searched using a set of parameters with guarantees of fully searching them. We introduce a new algorithm that searches the entire solution space using dynamic programming with a branch-and-bound technique both focused on solution subspaces. We devise a preprocessing phase that automatically tunes all of the parameters of the algorithm. This involves analyzing the results of combining sets of parameters on the entire solution space to come up with the best set of parameters. A bit more formally, ``the best set of parameters'' here means the best sets of coalitions to consider to search the entire solution space with the shortest run time. With experiments over several common value distributions, we show that the high-performing parameters selected by the preprocessing phase enable our algorithm to rapidly search the solution subspaces and outperform current state-of-the-art for several value distributions.

\end{abstract}


\section{Introduction}

\textit{Coalition formation} is an important research area in multiagent systems where several agents form coalitions to perform tasks. One of the key challenges that come along with this coalition formation is that of the coalition structure generation problem, which involves identifying the optimal set of coalitions between agents that maximizes the social welfare--- that is, an optimal coalition structure. Many problems require forming coalitions of agents. For instance, forming coalitions of delivery companies to reduce transportation costs by sharing deliveries~\cite{sandholm1997coalitions}, coalitions of human responders to coordinate their evacuation and rescue actions for disaster response~\cite{wu2020monte}, or grouping sensors and assigning them to different targets so that they provide better target coverage than they can individually~\cite{dang2006overlapping}.

To date, many algorithms have been developed for the coalition structure generation problem to address it in an optimal or approximate manner. These approaches range from dynamic programming~\cite{yeh1986dynamic,rahwan2008improved,michalak2016hybrid,changder2019effective}, which guarantee finding the optimal coalition structure but must be run to completion to do so, to branch-and-bound and graph-based search~\cite{sandholm1999coalition,rahwan2009anytime,taguelmimt2025aaai}, which provide intermediate solutions during the search, to heuristic algorithms~\cite{sen2000searching,keinanen2009simulated,di2010coalition,taguelmimt2021code,taguelmimt2022ictai}, which focus on speed and do not guarantee finding an optimal solution. The hybrid algorithms~\cite{michalak2016hybrid,Changder_Aknine_Ramchurn_Dutta_2021,taguelmimt2023ijcai,taguelmimt2024ijcai} such as ODP-IP~\cite{michalak2016hybrid} and BOSS~\cite{Changder_Aknine_Ramchurn_Dutta_2021}, which combine IDP~\cite{rahwan2008improved} and IP~\cite{rahwan2009anytime}, are the fastest exact algorithms for the CSG problem. However, as we show in this paper, the dynamic programming approaches used by these algorithms uses parameters that are hand-tuned, which negatively affects the performance of the resulting algorithms that are highly dependent of these parameters. 
Hand-tuning dynamic programming, while insuring that an optimal solution is found, as done in IDP (which is used by ODP-IP and BOSS) by only evaluating the coalitions of sizes $s \in \{2,..,\frac{2n}{3},n\}$ or in EDP~\cite{changder2019effective} by evaluating the coalition of even-numbered sizes is often 
not optimized. 

In light of this, and to make it possible to find high-performing parameter settings and optimize the algorithm configuration, we develop and present two algorithms that are focused on solution subspaces to optimize the search. In short, the main contributions of the paper are:

\begin{itemize}
    \item We develop a preprocessing phase that selects the best sets of parameters for dynamic programming. This enables a new kind of algorithm configuration approach for coalition structure generation.
    \item We introduce a new algorithm that searches the entire solution space using dynamic programming with a branch-and-bound technique both focused on solution subspaces. This leads to the first algorithm that is fully configured automatically by a preprocessing phase.
    \item We establish a baseline for future algorithm configuration in coalition structure generation research by providing initial theory.
    \item We empirically show that our algorithm outperforms current state-of-the-art for a number of value distributions when generating optimal solutions.
\end{itemize}

\section{Problem Formulation and Preliminaries}

For the remainder of this paper, we use the terms coalition structure and solution interchangeably. We study the \textit{coalition structure generation} (CSG) problem. A CSG problem instance is specified by a set $\mathcal{A} = \{a_{1}, a_{2}, . . . , a_{n}\}$ of $n$ agents, which is the size of the problem, and the value $v(\mathcal{C})$ of each coalition. The size of the CSG problem is $n$. A coalition $\mathcal{C}$ in $\mathcal{A}$ is any non-empty subset of $\mathcal{A}$. The size of $\mathcal{C}$ is $|\mathcal{C}|$. A characteristic function $v$ assigns a real value to each coalition. This value determines the efficiency of the coalition. A coalition structure $\mathcal{CS}$ is a partition of the set of agents $\mathcal{A}$ into disjoint coalitions. 
Formally, given a set of non-empty coalitions $\{\mathcal{C}_{1}, \mathcal{C}_{2},...,\mathcal{C}_{k}\}$, a coalition structure is a collection of coalitions $\mathcal{CS} =\{\mathcal{C}_{1}, \mathcal{C}_{2},...,\mathcal{C}_{k}\}$, where $k= |\mathcal{CS}|$, which satisfies the following constraints:  $\bigcup_{j=1}^{k} \mathcal{C}_{i} = \mathcal{A}$ and for all $i,j \in \{1,2,...,k\}$ where $i \neq j$, $\mathcal{C}_{i} \cap \mathcal{C}_{j} = \emptyset$.
\par
$\Pi(\mathcal{A})$ denotes the set of all coalition structures. The value of a coalition structure $\mathcal{CS}$ is assessed as the sum of the values of the disjoint coalitions that comprise it: $V (\mathcal{CS}) = \sum_{\mathcal{C} \in \mathcal{CS}} v(\mathcal{C})$. 
The coalition structure generation problem takes as input a set of agents $\mathcal{A}$ and aims at finding the optimal solution, which is the most valuable coalition structure $\mathcal{CS}^{*} \in \Pi(\mathcal{A})$, i.e., $\mathcal{CS}^{*}= \argmax_{\mathcal{CS} \in \Pi(\mathcal{A})} V(\mathcal{CS})$.

\begin{figure}[h!]
\centering
\resizebox{1.0\columnwidth}{4cm}{%
\begin{tikzpicture}
\large
\node[draw,rectangle,rounded corners=3pt] (a)at(7.5,0){$[1,1,1,1]$};

\node[draw,rectangle,rounded corners=3pt] (a)at(7.5,-2){$[1,1,2]$};

\node[draw,rectangle,rounded corners=3pt] (a)at(5.5,-4){$[1,3]$};

\node[draw,rectangle,rounded corners=3pt] (a)at(9.5,-4){$[2,2]$};

\node[draw,rectangle,rounded corners=3pt] (a)at(7.5,-6){$[4]$};

\node[] (a)at(-1.5,-6){$L_{1}: $};
\draw[>=latex,black!70,dashed,very thick] (7.0,-6) -- (-1.0,-6);
\node[] (a)at(-1.5,-4){$L_{2}: $};
\draw[>=latex,black!70,dashed,very thick] (0.1,-4) -- (-1.0,-4);
\node[] (a)at(-1.5,-2){$L_{3}: $};
\draw[>=latex,black!70,dashed,very thick] (6.5,-2) -- (-1.0,-2);
\node[] (a)at(-1.5,0){$L_{4}: $};
\draw[>=latex,black!70,dashed,very thick] (6.25,0) -- (-1.0,0);

\node[] (a)at(9.5,-6){$\Pi_{[4]}: $};
\node[] (a)at(11.3,-6){$\{\{a_{1},a_{2},a_{3},a_{4}\}\}$};

\node[] (a)at(10.8,-4){$\Pi_{[2,2]}: $};
\node[] (a)at(13,-4){$\{\{a_{1},a_{2}\},\{a_{3},a_{4}\}\}$};
\node[] (a)at(13,-4.5){$\{\{a_{1},a_{3}\},\{a_{2},a_{4}\}\}$};
\node[] (a)at(13,-5){$\{\{a_{1},a_{4}\},\{a_{2},a_{3}\}\}$};

\node[] (a)at(0.8,-4){$\Pi_{[1,3]}: $};
\node[] (a)at(3,-4){$\{\{a_{1}\},\{a_{2},a_{3},a_{4}\}\}$};
\node[] (a)at(3,-4.5){$\{\{a_{2}\},\{a_{1},a_{3},a_{4}\}\}$};
\node[] (a)at(3,-5){$\{\{a_{3}\},\{a_{1},a_{2},a_{4}\}\}$};
\node[] (a)at(3,-5.5){$\{\{a_{4}\},\{a_{1},a_{2},a_{3}\}\}$};

\node[] (a)at(9.0,-2){$\Pi_{[1,1,2]}: $};
\node[] (a)at(11.5,-2){$\{\{a_{1}\},\{a_{2}\},\{a_{3},a_{4}\}\}$};
\node[] (a)at(11.5,-2.5){$\{\{a_{1}\},\{a_{3}\},\{a_{2},a_{4}\}\}$};
\node[] (a)at(11.5,-3){$\{\{a_{1}\},\{a_{4}\},\{a_{2},a_{3}\}\}$};
\node[] (a)at(15.3,-2){$\{\{a_{2}\},\{a_{3}\},\{a_{1},a_{4}\}\}$};
\node[] (a)at(15.3,-2.5){$\{\{a_{2}\},\{a_{4}\},\{a_{1},a_{3}\}\}$};
\node[] (a)at(15.3,-3){$\{\{a_{3}\},\{a_{4}\},\{a_{1},a_{2}\}\}$};

\node[] (a)at(13.4,-2){$,$};
\node[] (a)at(13.4,-2.5){$,$};
\node[] (a)at(13.4,-3){$,$};

\node[] (a)at(9.5,0){$\Pi_{[1,1,1,1]}: $};
\node[] (a)at(12.3,0){{$\{\{a_{1}\},\{a_{2}\},\{a_{3}\},\{a_{4}\}\}$}};

\node[] (a)at(9.5,-5.3){\small$4=2+2$};
\node[] (a)at(9.5,-4.9){\small Split $4$};
\node[] (a)at(5.5,-5.3){\small$4=1+3$};
\node[] (a)at(5.5,-4.9){\small Split $4$};

\node[] (a)at(5.2,-3.3){\small$3=1+2$};
\node[] (a)at(5.2,-2.9){\small Split $3$};
\node[] (a)at(7.95,-3.5){\small$2=1+1$};
\node[] (a)at(7.95,-3.1){\small Split $2$};

\node[] (a)at(6.7,-1.2){\small$2=1+1$};
\node[] (a)at(6.7,-0.8){\small Split $2$};

\draw[->,>=latex,LightCyan2!95,very thick] (7.5,-5.7) -- (5.5,-4.3);
\draw[->,>=latex,vert,very thick] (7.5,-5.7) -- (9.5,-4.3);
\draw[->,>=latex,red,dashed,very thick] (5.5,-3.7) -- (7.4,-2.3);
\draw[->,>=latex,vert,very thick] (9.5,-3.7) -- (7.5,-2.3);
\draw[->,>=latex,LightCyan2!95,thick] (7.5,-1.7) -- (7.5,-0.3);

\end{tikzpicture}%
}
\caption{A four-agent integer partition graph of 4 levels of nodes.} \label{IPG}
\end{figure}
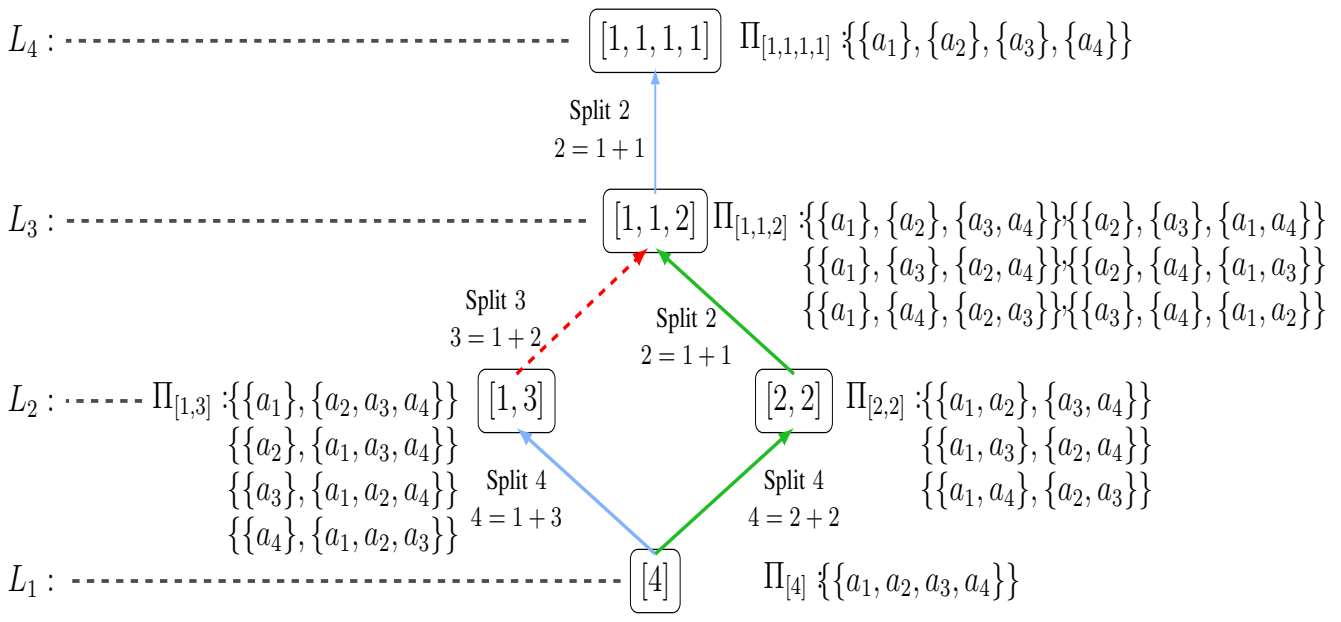

The integer partition graph~\cite{rahwan2009anytime} is a representation that divides the space of all the coalition structures into subspaces that are each represented by an integer partition of $n$. For instance, for $n=4$ agents, the integer partitions are: $[4],[1,3],[2,2],[1,1,2],[1,1,1,1]$. Given $n$ agents, with this representation, each integer partition of $n$ is represented by a node, where two adjacent nodes are connected if and only if the integer partition in level $i$ can be obtained from the one in level $i-1$ by splitting only an integer. Fig. \ref{IPG} shows a four-agent example.
\par
Each level $i \in \{1,2,..,n\}$ in the integer partition graph contains nodes representing integer partitions of $n$ containing $i$ parts. For instance, level 2 contains nodes where integer partitions of $n$ have two parts (see Fig. \ref{IPG}). In the integer partition graph, each integer partition $\mathcal{P}$ represents a subspace of solutions that contains a set of coalition structures in which the sizes of the coalitions match the parts of $\mathcal{P}$. For example, the node [1,1,2] consists of all coalition structures that contain two coalitions of size 1 and one coalition of size 2. A four agent example is shown in Fig. \ref{IPG} where all the coalition structures corresponding to each node in the integer partition graph are shown.

\subsection{Related Work}

To date, a number of approaches have been proposed to solve the CSG problem either optimally or approximately. Many works~\cite{yeh1986dynamic,rahwan2008improved,michalak2016hybrid} proposed \textit{dynamic programming} algorithms which guarantee finding the optimal coalition structure but must be run to completion to do so. To allow returning solutions prior to termination, a number of works~\cite{sandholm1999coalition,dang2004generating,rahwan2009anytime} have turned to developing \textit{anytime} algorithms in an effort to allow premature termination while providing intermediate solutions at each point during the execution. When the number of agents increases and the problem becomes too hard, \textit{heuristic} algorithms present the only practical option. For this reason, many works~\cite{sen2000searching,keinanen2009simulated,di2010coalition} have presented such algorithms, which focus on speed and do not guarantee finding an optimal solution. Till today, the hybrid algorithms ODP-IP and BOSS, which combine IDP~\cite{rahwan2008improved} and IP~\cite{rahwan2009anytime}, are the fastest exact algorithms for the CSG problem.


\section{Subspace-Focused Search Method}

Our method consists of a preprocessing phase and a search phase. The preprocessing phase is used to provide the algorithm that we use in the search phase with the best set of parameters that include the number of processes and the sets of coalitions to evaluate in each process. The search phase uses the \textit{SEFAR (Subspace-oriEnted for Focused dynAmic pRogramming)} algorithm, which combines dynamic programming with solution subspace pruning optimization for finding optimal solutions. The algorithm thus evaluates sets of coalitions selected by the preprocessing phase.

\subsection{Preprocessing Phase}

The preprocessing phase computes the best sets of coalitions to evaluate in the search phase by the SEFAR algorithm. 
To select the coalitions to evaluate, the preprocessor proceeds in several iteration and returns, after each iteration, a set of coalition sizes $\mathcal{S}$ to consider for evaluating the coalitions. This set of coalition sizes $\mathcal{S}$ 
identifies all coalitions of size $|\mathcal{C}| \in \mathcal{S}$. Before going any further, let us introduce the fundamental principle used by the preprocessor. To fully search a solution subspace, not all coalitions need to be evaluated. In fact, several sets of coalitions can be considered for evaluation to fully search a subspace. This can be particularly visualized in the integer partition graph (see Fig. \ref{IPG}).

Starting from the bottom node, each edge that connects nodes between them results from splitting a coalition size into two, which means that all coalitions of that size are evaluated. For example, the red edge that connects the node $[1,3]$ to the node $[1,1,2]$ reflects a splitting of the coalition size 3 into 1 and 2 ($3=1+2$), which means that all the coalitions of size 3 are evaluated (they are split into two coalitions of sizes 1 and 2 to obtain their best partitions). Hence, dividing a coalition into two corresponds to an upward movement in the graph (e.g. from $[1,3]$ to $[1,1,2]$ with a coalition of size 3, or from $[2,2]$ to $[1,1,2]$ and from $[1,1,2]$ to $[1,1,1,1]$ with a coalition of size 2). The following theorem implies that to fully search a subspace of solutions, it is not required to consider all coalition sizes.

\begin{thm}
By considering only a subset of the edges in the graph, if a node $\mathcal{N}$ in the integer partition graph is connected to the bottom node with a series of these edges, the node $\mathcal{N}$ will be fully searched.
\end{thm}

\begin{proof}
An edge in the integer partition graph represents a splitting of a coalition into two coalitions. A path between the bottom node and another node $\mathcal{N}$ represents the possibility of obtaining any coalition structure of that node through a series of recursive splittings of the coalitions from the grand coalition. Thus, if there is a path between the bottom node and $\mathcal{N}$, all the coalition structures of $\mathcal{N}$ will be generated, and hence $\mathcal{N}$ will be fully searched.
\end{proof}

With this in mind, there are several sets of edges that we can consider to connect a subspace of the integer partition graph to the bottom node and fully search it. Since an edge reflects a splitting of a coalition size, we can choose which coalition sizes (and hence which coalitions) to split to achieve this. For example, to fully search the subspace $[1,1,2]$, we only need one path to the bottom node. For instance, the green edges connect this node $[1,1,2]$ to the bottom node. Hence, considering only the coalition sizes 4 (to connect $[2,2]$ to $[4]$) and 2 (to connect $[1,1,2]$ to $[2,2]$) is sufficient. However, this is not the only path that can connect the subspace to the bottom node and hence $\{2,4\}$ is not the only set of coalition sizes to consider that can do so ($\{2,3\}$ and $\{2,3,4\}$ in our example can also connect $[1,1,2]$ to the bottom node). However, to each set of sizes corresponds a different run time, which corresponds to the time for evaluating all the coalitions of those sizes. As a result, there is an optimal set of sizes, which by evaluating the corresponding coalitions, we can achieve the shortest run time to fully search a node.

To find the optimal set of sizes for fully searching a node, our preprocessor starts by determining the estimated run time for evaluating the coalitions of each size (i.e. the cost of considering each size). It corresponds to the evaluation time of all the different ways of splitting the coalitions of each size. The estimated time of a splitting is the computational cost associated with this hardware operation which is a fixed value. Then, the preprocessor computes the optimal set of sizes that produces the shortest run time. The estimated time for a set of sizes is the sum of the evaluation times of the coalitions of the size set. To find the optimal set, it goes through each possible set of sizes, and for each possible set, it tests whether the node is connected to the bottom node by only considering the edges that result from splitting the sizes of the set. If so, the preprocessor tests whether the corresponding run time is better than the last best one and in that case the set becomes the best set. Once all possible sets are tested, the preprocessor obtains the optimal set. For example with the node $[2,2]$, the preprocessor finds that $\{4\}$, $\{2,4\}$, $\{3,4\}$ and $\{2,3,4\}$ as the only sets that guarantee fully searching the node $[2,2]$ and $\{2,4\}$ as the optimal one.

This process is repeated until the optimal set for each subspace is obtained. However, we do not do this for each subspace iteratively. Once the optimal set for a subspace is obtained, we eliminate all the subspaces that can be generated by this size set and proceed with the remaining ones. Notice that several subspaces can be generated by the same size set. For example, with the size set $\{4\}$ we generate the node $[2,2]$, but also the node $[1,3]$ and hence it is not necessary to compute the optimal set of sizes for the node $[1,3]$. This is why we do not do this for each subspace. At the end, the preprocessor returns sets of coalition sizes that search all the subspaces of the integer partition graph. 

We run the preprocessor independently of the search phase because its output is unique and is a configuration of the SEFAR algorithm. Therefore, it is executed only once. Algorithm \ref{prepro} shows how the preprocessor computes the sets of coalition sizes and Algorithm \ref{COS} shows how it computes the optimal set for one subspace. Here, a set of coalition sizes is represented in binary format. We use $n-2$ bits to represent every set, where each bit corresponds to a coalition size. We do not consider the coalition sizes 1 and $n$ because the size 1 coalitions are never split and the size $n$ coalition is always split. Hence, we consider $n-2$ coalition sizes from 2 to $n-1$.  The bits set to 1 in this number represent the selected coalition sizes of the corresponding set. For example with four agents, the possible sets of sizes are $\{2\}$, $\{3\}$, and $\{2,3\}$. These sets are represented by the binary numbers $01_{2}$, $10_{2}$, and $11_{2}$, respectively. Note that the coalition of size 4 is always retained. That is why it is not considered during the calculation, but is added at the end (see line 8).

\subsubsection{Reducing the Number of Sets}

Once all the sets of sizes are obtained by the preprocessor, we can work on them to reduce their number because some of these sets become not necessary at the end. To see why this additional processing is possible, suppose we are computing the sets of sizes but we are starting with a subspace easier to search and its size set is $\mathcal{S}_{1}$. Few iteration after that we find another set of sizes $\mathcal{S}_{2}$ for another subspace. If by considering the size set $\mathcal{S}_{2}$ we can generate all the subspaces generated with the set of sizes $\mathcal{S}_{1}$, the set $\mathcal{S}_{1}$ becomes not necessary as it does not generate a subspace that is not generated with another set. For example with 4 agents, if the preprocessor starts with the subspace $[2,2]$ and finds the set $\mathcal{S}_{1} = \{4\}$ to fully search it and then when it computes the optimal set for the node $[1,1,1,1]$, it finds the set $\mathcal{S}_{2} = \{2,4\}$. The set $\mathcal{S}_{2}$ generates all the subspaces including the subspaces $[2,2]$ and $[1,3]$ generated by $\mathcal{S}_{1}$. Hence, the set $\mathcal{S}_{1}$ becomes irrelevant.

By removing the unnecessary sets of coalition sizes, the preprocessor reduces the number of processes of the SEFAR algorithm. Note that the number of processes and the sets of coalitions to evaluate in each process are defined by the preprocessor and further refined after this step.

\subsubsection{Ordering the Subspaces}

Given the nodes of the integer partition graph, we can start searching for the sets of sizes from any node. The fact that we reduce the number of sets means that the order in which we consider the subspaces is not important as far as the preprocessor is concerned. Even though this would generate the optimal sets of sizes, not guiding the search for these sets would result in a larger number of iterations of the algorithm. Our goal is then to start with nodes which would require several edges (and hence sizes) to be considered to connect it to the bottom node. This way we minimize the likelihood of those unnecessary sets being considered before those that make them irrelevant. In our context, the nodes to consider first are the farthest from the bottom node, and thus the ones that belong to the last levels. In Fig. \ref{IPG} for example, if the preprocessor starts with the node $[1,1,1,1]$ in the first iteration, then it will find the set of sizes $\{2,4\}$ as the optimal set, prune all the other subspaces, and end after one iteration, thereby reducing the number of iterations and irrelevant sets. This is indeed plausible, as starting with any other subspace, except $[1,1,2]$, would end up generating more irrelevant sets and iterations. 

\begin{algorithm}[h!]
\DontPrintSemicolon
 Generate all the nodes of the integer partition graph and store them in $\mathcal{N}$.\;
 \If($\triangleright$ If there is still nodes that are not generated by the selected size sets){$\mathcal{N}$ is not empty}{
 Select one node from $\mathcal{N}$ and compute the best size set $\mathcal{S}$ that generates it with the shortest run time.\;
 Add $\mathcal{S}$ to the set of the best size sets $\mathcal{SZ}$.\;
 Remove from $\mathcal{N}$ all the nodes that are also generated with the size set $\mathcal{S}$.\;
 Go to line 2 and start with the remaining nodes.\;
 }
 \nonl $\triangleright$ At this stage, the preprocessor computed the best size sets, which when combined, allow to search the entire solution space. However, some of these sets may be unnecessary. This can happen if a size set was considered before the computation of the set that makes it become unnecessary. Hence, the preprocessor removes them.\;
 Remove from $\mathcal{SZ}$ the size sets that are subsets of others.\;
 Return $\mathcal{SZ}$\;
 \caption{Preprocessor for computing the coalition sets} \label{prepro}
\end{algorithm}

\begin{algorithm}[h!]
\KwIn{A subspace $\mathcal{N}$, a CSG problem size $n$, and the times to split the coalitions of sizes 2 to $n$-1.}
\KwOut{The best set of coalition sizes $\mathcal{S}$ to fully search the subspace $\mathcal{N}$. $\mathcal{S}$ is in binary format.}
\DontPrintSemicolon
 $\mathcal{S} \leftarrow 2^{n-2}-1$ $\triangleright$ Initially, the best set includes all the coalition sizes, i.e. $\{2,..,n-1\}$. $2^{n-2}-1$ is the binary representation of this set.\;
 $t^{*} \leftarrow SetTime(2^{n-2}-1)$ $\triangleright$ $SetTime$ returns the run time for a set of sizes. $t^{*}$ is initialized to the run time when considering all the coalition sizes, which is the worst time.\;
 \For($\triangleright$ $i$ corresponds to a set of coalition sizes represented in binary format.){$i= 1$ to $2^{n-2}-1$}{
 $x \leftarrow GeneratedSubspaces(i)$\;
 \If($\triangleright$ the set $i$ generates the subspace $\mathcal{N}$.){$\mathcal{N} \in x$ \textbf{and} $SetTime(i) < t^{*}$}{
 $\mathcal{S} \leftarrow i$\;
 $t^{*} \leftarrow SetTime(i)$\;
 }
 }
 Add $n$ to $\mathcal{S}$ $\triangleright$ $n$ is always considered.\;
 Return $\mathcal{S}$\;
 \caption{Computing Optimal Set} \label{COS}
\end{algorithm}

\subsection{Search Phase: SEFAR Algorithm}

To search for the optimal solution, the SEFAR algorithm uses multiple parallel processes, each with a set of coalition sizes as input with which it explores a certain number of subspaces. The number of processes used by the algorithm is defined by the preprocessor and corresponds to the number of unique size sets found. Hence, the resulting SEFAR algorithm may have different numbers of processes for different problem sizes. The highest valued coalition structure returned by these processes is an optimal coalition structure.

Each process is based on dynamic programming and receives a set of sizes and a partial integer partition graph with all subspaces and no edges as input. Every process evaluates the selected coalitions by the set of sizes and computes the best partition for each coalition and stores it in a partition table $\mathcal{P}_{t}$ along with its value in a value table $\mathcal{V}_{t}$. 

In each process of SEFAR, the search is carried out on the coalitions of sizes that belong to the set of sizes $\mathcal{S} = \{s_{1},s_{2},...,s_{k}\}$ that it received as input, which is obtained by the preprocessor. SEFAR starts by evaluating the coalitions of size $s_{1}$, and for every coalition $\mathcal{C}_{s_{1}}$ of size $s_{1}$ that the algorithm encounters, it evaluates all possible ways of splitting $\mathcal{C}_{s_{1}}$ into two coalitions and checks whether it is beneficial to split $\mathcal{C}_{s_{1}}$ or not. The best results are then stored in $\mathcal{P}_{t}(\mathcal{C}_{s_{1}})$ and $\mathcal{V}_{t}(\mathcal{C}_{s_{1}})$. Similarly, SEFAR evaluates the coalitions of size $s_{2}$, and so on. This process is repeated until the last size $s_{k}$. This way, by the time RDP evaluates the coalitions of size $s_{k}$, which is always equal to $n$ (because we always need to compute the best partition for the coalition of size $n$ to find the optimal solution), the best partition of each coalition is obtained, and also that of the grand coalition that corresponds to the optimal solution. 

Often, some subspaces have no chance of containing an optimal solution. To enable pruning out these subspaces, we introduce the upper bound $UB(\mathcal{N})$ of a subspace $\mathcal{N}$, which is the highest value a coalition structure of this subspace can possibly reach. To this end, let $Max_{s}$ denote the highest value of the coalitions of size $s$. By computing $Max_{s}$ for every coalition size $s$, it is possible to compute the upper bound of a subspace $\mathcal{S}$: $UB_{\mathcal{S}} = \sum_{s \in Integers(\mathcal{S})} Max_{s}$, where $Integers(\mathcal{S})$ is the set of integers that form the integer partition of the subspace $\mathcal{S}$. For example, for $\mathcal{S} = [1,2,3,4]$, $Integers(\mathcal{S}) = \{1,2,3,4\}$ and the upper bound of $\mathcal{S}$ is $UB_{[1,2,3,4]} = \sum_{s \in \{1,2,3,4\}} Max_{s} = Max_{1} + Max_{2} + Max_{3} + Max_{4}$. By comparing the upper bounds of the subspaces, the processes identify those that have no chance of improving the solution quality and prune the corresponding nodes that do not have a better upper bound than the last best solution found. Moreover, after a process evaluates all the coalitions of size $x$, it adds the edges that result from splitting $x$ into two integers to the integer partition graph. These newly added edges make a certain number of subspaces become reachable from the bottom node, meaning that they are fully searched, thereby pruning them out. 
Also, each time a process finishes evaluating the coalitions of a size, the subspace pruning, using the upper bounds and the connection to the bottom node is repeated. Hence, the SEFAR algorithm finishes and returns the optimal solution when all subspaces are pruned. 

The pseudo code of SEFAR is detailed in Algorithm \ref{sefaralg}.

\begin{algorithm}[h!]
\KwIn{Set of all possible coalitions and the value $V_{t}(\mathcal{C})$ and $P_{t}(\mathcal{C})$ of each coalition $\mathcal{C}$. A number of agents $n$. Sets of coalition sizes $\mathcal{S}_{i}$ to consider by SEFAR.}
\KwOut{An optimal coalition structure $\mathcal{CS}^{*}$ and its value.}
\DontPrintSemicolon
\For($\triangleright$ $|\mathcal{S}_{i}|$ is the number of considered sets of sizes.){$i = 1$ \textbf{to} $|\mathcal{S}_{i}|$}{
 Generate a partial integer partition graph $\mathcal{I}_{\mathcal{G}}$ with all subspaces and with no edges\;
 \nonl $\triangleright$ Begin parallel\;
 \nonl $\triangleright$ SEFAR process runs in parallel given $\mathcal{S}_{i}$\;
 $\mathcal{CS}^{*}$, $\mathcal{V}^{*}$ $\leftarrow$ $Search\_Process(\mathcal{P}_{t},\mathcal{V}_{t},n,\mathcal{S}_{i},\mathcal{I}_{\mathcal{G}})$\;
 \nonl $\triangleright$ End parallel\;
}
 Return $\mathcal{CS}^{*}$, $\mathcal{V}^{*}$\;
 \caption{SEFAR algorithm} \label{sefaralg}
\end{algorithm}

\begin{algorithm}[h!]
\KwIn{The partition and value tables $\mathcal{P}_{t}$ and $\mathcal{V}_{t}$. A number of agents $n$. Set of coalition sizes $\mathcal{S}$ to consider by the process. A partial integer partition graph.}
\KwOut{An optimal coalition structure $CS^{*}$ and its value.}
\DontPrintSemicolon
 \For{$s \in \mathcal{S}$}{
\ForEach{$C \subseteq A$, where $|C| = s$}{
\ForEach{$C_{1}, C_{2} \subseteq C$, where $C_{1} \cup C_{2} = C$ and $C_{1} \cap C_{2} = \emptyset$}{
\If{$V_{t}(C_{1}) + V_{t}(C_{2}) > V_{t}(C)$}{
$V_{t}(C) \leftarrow V_{t}(C_{1}) + V_{t}(C_{2})$\;
$P_{t}(C) \leftarrow \{C_{1},C_{2}\}$\;
}
}
}
Compute the best solution, {\scriptsize$\mathcal{CS}^{*}$, $\mathcal{V}^{*}$}, from $P_{t}$ and $V_{t}$\;
Add to the integer partition graph the edges that result from splitting the size $s$\;
Prune the subspaces connected to the bottom node $\triangleright$ these subspaces are already explored\;
Prune the subspaces that do not have a better upper bound than the last best solution found\;
\If{all the subspaces are pruned}{
Return $\mathcal{CS}^{*}$, $\mathcal{V}^{*}$\;
}
}
 \caption{Search Process} \label{gradp}
\end{algorithm}

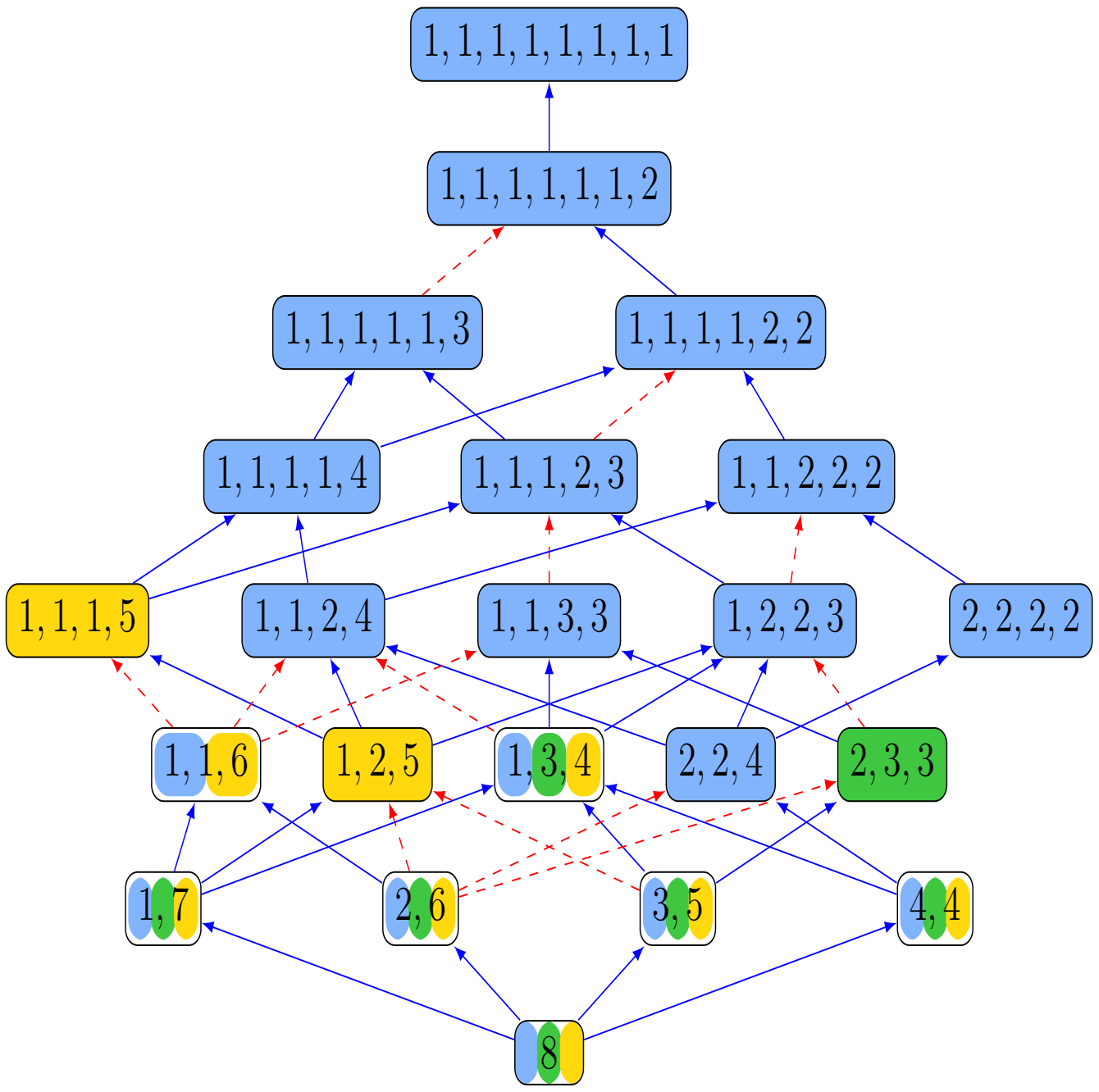
\begin{figure}[h!]
\centering
\resizebox{0.99\columnwidth}{8.7cm}{%
\begin{tikzpicture}

\large

\node[draw,rectangle,rounded corners=4pt, fill=LightCyan2!95] (a22)at(7.5,0){$1,1,1,1,1,1,1,1$};

\node[draw,rectangle,rounded corners=4pt, fill=LightCyan2!95] (a21)at(7.5,-1.25){$1,1,1,1,1,1,2$};

\node[draw,rectangle,rounded corners=4pt, fill=LightCyan2!95] (a20)at(5.5,-2.5){$1,1,1,1,1,3$};

\node[draw,rectangle,rounded corners=4pt, fill=LightCyan2!95] (a19)at(9.5,-2.5){$1,1,1,1,2,2$};

\node[draw,rectangle,rounded corners=4pt, fill=LightCyan2!95] (a18)at(4.5,-3.75){$1,1,1,1,4$};

\node[draw,rectangle,rounded corners=4pt, fill=LightCyan2!95] (a17)at(7.5,-3.75){$1,1,1,2,3$};

\node[draw,rectangle,rounded corners=4pt, fill=LightCyan2!95] (a16)at(10.5,-3.75){$1,1,2,2,2$};

\node[draw,rectangle,rounded corners=4pt, fill=gold!95] (a15)at(2,-5){$1,1,1,5$};

\node[draw,rectangle,rounded corners=4pt,fill=LightCyan2!95] (a14)at(4.75,-5){$1,1,2,4$};

\node[draw,rectangle,rounded corners=4pt, fill=LightCyan2!95] (a13)at(7.5,-5){$1,1,3,3$};

\node[draw,rectangle,rounded corners=4pt,fill=LightCyan2!95] (a12)at(10.25,-5){$1,2,2,3$};

\node[draw,rectangle,rounded corners=4pt, fill=LightCyan2!95] (a11)at(13,-5){$2,2,2,2$};

\node[rectangle,fill=LightCyan2!95,rounded corners=5pt, minimum width=0.6cm, minimum height=0.55cm, align=center] (a) at (3.2,-6.25) {};
\node[rectangle,fill=gold!95,rounded corners=5pt, minimum width=0.6cm, minimum height=0.55cm, align=center] (a) at (3.8,-6.25) {};
\node[draw,rectangle,rounded corners=4pt] (a10)at(3.5,-6.25){$1,1,6$};

\node[draw,rectangle,rounded corners=4pt,fill=gold!95] (a9)at(5.5,-6.25){$1,2,5$};

\node[rectangle,fill=LightCyan2!95,rounded corners=5pt, minimum width=0.4cm, minimum height=0.55cm, align=center] (a) at (7.1,-6.25) {};
\node[rectangle,fill=vert!85,rounded corners=5pt, minimum width=0.4cm, minimum height=0.55cm, align=center] (a) at (7.5,-6.25) {};
\node[rectangle,fill=gold!95,rounded corners=5pt, minimum width=0.4cm, minimum height=0.55cm, align=center] (a) at (7.9,-6.25) {};
\node[draw,rectangle,rounded corners=4pt] (a8)at(7.5,-6.25){$1,3,4$};

\node[draw,rectangle,rounded corners=4pt, fill=LightCyan2!95] (a7)at(9.5,-6.25){$2,2,4$};

\node[draw,rectangle,rounded corners=4pt, fill=vert!85] (a6)at(11.5,-6.25){$2,3,3$};

\node[rectangle,fill=LightCyan2!95,rounded corners=5pt, minimum width=0.266cm, minimum height=0.55cm, align=center] (a) at (2.733,-7.5) {};
\node[rectangle,fill=vert!85,rounded corners=5pt, minimum width=0.266cm, minimum height=0.55cm, align=center] (a) at (3.0,-7.5) {};
\node[rectangle,fill=gold!95,rounded corners=5pt, minimum width=0.266cm, minimum height=0.55cm, align=center] (a) at (3.266,-7.5) {};
\node[draw,rectangle,rounded corners=4pt] (a5)at(3.0,-7.5){$1,7$};

\node[rectangle,fill=LightCyan2!95,rounded corners=5pt, minimum width=0.266cm, minimum height=0.55cm, align=center] (a) at (5.733,-7.5) {};
\node[rectangle,fill=vert!85,rounded corners=5pt, minimum width=0.266cm, minimum height=0.55cm, align=center] (a) at (6.0,-7.5) {};
\node[rectangle,fill=gold!95,rounded corners=5pt, minimum width=0.266cm, minimum height=0.55cm, align=center] (a) at (6.266,-7.5) {};
\node[draw,rectangle,rounded corners=4pt] (a4)at(6,-7.5){$2,6$};

\node[rectangle,fill=LightCyan2!95,rounded corners=5pt, minimum width=0.266cm, minimum height=0.55cm, align=center] (a) at (8.733,-7.5) {};
\node[rectangle,fill=vert!85,rounded corners=5pt, minimum width=0.266cm, minimum height=0.55cm, align=center] (a) at (9.0,-7.5) {};
\node[rectangle,fill=gold!95,rounded corners=5pt, minimum width=0.266cm, minimum height=0.55cm, align=center] (a) at (9.266,-7.5) {};
\node[draw,rectangle,rounded corners=4pt] (a3)at(9,-7.5){$3,5$};

\node[rectangle,fill=LightCyan2!95,rounded corners=5pt, minimum width=0.266cm, minimum height=0.55cm, align=center] (a) at (11.733,-7.5) {};
\node[rectangle,fill=vert!85,rounded corners=5pt, minimum width=0.266cm, minimum height=0.55cm, align=center] (a) at (12.0,-7.5) {};
\node[rectangle,fill=gold!95,rounded corners=5pt, minimum width=0.266cm, minimum height=0.55cm, align=center] (a) at (12.266,-7.5) {};
\node[draw,rectangle,rounded corners=4pt] (a2)at(12.0,-7.5){$4,4$};

\node[rectangle,fill=LightCyan2!95,rounded corners=5pt, minimum width=0.266cm, minimum height=0.55cm, align=center] (a) at (7.233,-8.75) {};
\node[rectangle,fill=vert!85,rounded corners=5pt, minimum width=0.266cm, minimum height=0.55cm, align=center] (a) at (7.5,-8.75) {};
\node[rectangle,fill=gold!95,rounded corners=5pt, minimum width=0.266cm, minimum height=0.55cm, align=center] (a) at (7.766,-8.75) {};
\node[draw,rectangle, minimum width=0.8cm,rounded corners=4pt] (a1)at(7.5,-8.75){$8$};

\draw[->,>=latex,blue] (a1) -- (a2);
\draw[->,>=latex,blue] (a1) -- (a3);
\draw[->,>=latex,blue] (a1) -- (a4);
\draw[->,>=latex,blue] (a1) -- (a5);

\draw[->,>=latex,blue] (a2) -- (a7);
\draw[->,>=latex,blue] (a2) -- (a8);

\draw[->,>=latex,blue] (a3) -- (a6);
\draw[->,>=latex,blue] (a3) -- (a8);
\draw[->,>=latex,red,dashed] (a3) -- (a9);

\draw[->,>=latex,red,dashed] (a4) -- (a6);
\draw[->,>=latex,red,dashed] (a4) -- (a7);
\draw[->,>=latex,red,dashed] (a4) -- (a9);
\draw[->,>=latex,blue] (a4) -- (a10);

\draw[->,>=latex,blue] (a5) -- (a8);
\draw[->,>=latex,blue] (a5) -- (a9);
\draw[->,>=latex,blue] (a5) -- (a10);

\draw[->,>=latex,red,dashed] (a6) -- (a12);
\draw[->,>=latex,blue] (a6) -- (a13);

\draw[->,>=latex,blue] (a7) -- (a11);
\draw[->,>=latex,blue] (a7) -- (a12);
\draw[->,>=latex,blue] (a7) -- (a14);

\draw[->,>=latex,blue] (a8) -- (a12);
\draw[->,>=latex,blue] (a8) -- (a13);
\draw[->,>=latex,red,dashed] (a8) -- (a14);

\draw[->,>=latex,blue] (a9) -- (a12);
\draw[->,>=latex,blue] (a9) -- (a14);
\draw[->,>=latex,blue] (a9) -- (a15);

\draw[->,>=latex,red,dashed] (a10) -- (a13);
\draw[->,>=latex,red,dashed] (a10) -- (a14);
\draw[->,>=latex,red,dashed] (a10) -- (a15);

\draw[->,>=latex,blue] (a11) -- (a16);

\draw[->,>=latex,red,dashed] (a12) -- (a16);
\draw[->,>=latex,blue] (a12) -- (a17);

\draw[->,>=latex,red,dashed] (a13) -- (a17);

\draw[->,>=latex,blue] (a14) -- (a16);
\draw[->,>=latex,blue] (a14) -- (a18);

\draw[->,>=latex,blue] (a15) -- (a17);
\draw[->,>=latex,blue] (a15) -- (a18);

\draw[->,>=latex,blue] (a16) -- (a19);

\draw[->,>=latex,red,dashed] (a17) -- (a19);
\draw[->,>=latex,blue] (a17) -- (a20);

\draw[->,>=latex,blue] (a18) -- (a19);
\draw[->,>=latex,blue] (a18) -- (a20);

\draw[->,>=latex,blue] (a19) -- (a21);

\draw[->,>=latex,red,dashed] (a20) -- (a21);

\draw[->,>=latex,blue] (a21) -- (a22);

\end{tikzpicture}%
}
\caption{Illustration of SEFAR on a 8-agent integer partition graph. SEFAR uses three processes to evaluate the coalitions of sizes $s \in \mathcal{S}_{1} = \{2,7,8\}$ in the first process, the coalitions of sizes $s \in \mathcal{S}_{2} = \{5,8\}$ in the second process and the coalitions of sizes $s \in \mathcal{S}_{3} = \{2,4,8\}$ in the third process. After finishing the evaluation process, SEFAR will have fully searched, with the set $\mathcal{S}_{1}$, all the yellow subspaces, with the set $\mathcal{S}_{2}$, all the green subspaces, and with the set $\mathcal{S}_{3}$, all the blue subspaces. Hence, every subspace of the integer partition graph will be fully searched by at least one process.} \label{IPGSEFAR}
\end{figure}

Fig. \ref{IPGSEFAR} shows an illustration example of the working of SEFAR with 8 agents. For this example, the preprocessor finds three sets of sizes to search the subspaces with the shortest run time, namely $\mathcal{S}_{1} = \{2,7,8\}$, $\mathcal{S}_{2} = \{5,8\}$ and $\mathcal{S}_{3} = \{2,4,8\}$. This way, each subspace will be fully searched using at least one set of sizes (i.e. by at least one process). The yellow nodes represent the subspaces searched using $\mathcal{S}_{1}$, the green nodes represent the subspaces searched using $\mathcal{S}_{2}$ and the blue nodes represent the subspaces searched using $\mathcal{S}_{3}$. For example, the node $[1,3,4]$ contains the three colors and hence, it is fully searched by the three processes. The node $[1,1,6]$ contains the yellow and blue colors, meaning that it is fully searched by the first and third processes. Some nodes contain only one color, meaning that they are fully searched by only one process. For instance, the node $[2,3,3]$ contains only the green color and hence, it is only searched by the second process using $\mathcal{S}_{2}$. The edges in Fig. \ref{IPGSEFAR} are of two kinds. The blue edges represent the splitting of sizes considered by the processes, i.e. $2,4,5,7,8$, while the red edges represent the skipping of sizes not considered by the processes, i.e. $3,6$.

\subsubsection{Extension: Reducing Memory Requirements}

We extend the algorithm from the previous subsection to avoid storing the $V_{t}$ table for each SEFAR process. To save memory, we store the $V_{t}$ table only once and use it for all processes of SEFAR. Observe that for each SEFAR process, we need to compute the value $V_{t}(\mathcal{C})$ for all considered coalitions in the set of sizes during each execution of the algorithm. However, given that the processes of the SEFAR algorithm gradually store the maximum values of the coalitions, working with the same $V_{t}$ table does not interfere with this working. The underling reason is that for each process of SEFAR, the $V_{t}$ table stores at least the needed values. In case the value of a coalition is higher than the one computed, it only enables the SEFAR process to search more coalition structures and never less. Based on this, the following theorem holds.

\begin{thm}
Let $k$ be the number of processes used by SEFAR for $n$ agents. SEFAR avoids storing $(k-1) \times (2^{n}$-$1)$ coalition values.
\end{thm}

\begin{proof}
Fix a number of agents $n$ and a process of SEFAR. To find the optimal solution in the searched subspaces, a process needs to store the values of all the $2^{n}$-$1$ coalitions (even if many of them are not evaluated because they are used to evaluate other coalitions). This results in storing the coalition values $k$ times. By only storing the coalition values once, we avoid storing $k-1$ times these values. Hence, SEFAR avoids storing $(k-1) \times (2^{n}$-$1)$ coalition values.
\end{proof}

\section{Analysis of the Algorithms}

In this section, we prove that the proposed algorithm is exact and analyze the computational complexity of the SEFAR algorithm and the preprocessor. Recall that the preprocessor is run only once for each problem size $n$ to configure the algorithm. Hence, SEFAR is the algorithm we run to search for the optimal solutions.

\begin{thm}
The SEFAR algorithm always finds the optimal solution.
\end{thm}
\begin{proof}
Each node in the integer partition graph is searched using SEFAR by at least one process with one set of sizes. A node represents a subspace, which contains a number of coalition structures that match the parts of the subspace. The subspaces are shared, by the preprocessor, between the processes of SEFAR so that each subspace is fully searched by at least one process. Hence, once the processes finish evaluating the coalitions, all the subspaces will be fully searched. Thus, the optimal solution is found.
\end{proof}

\begin{thm}
Given $n$ agents, SEFAR runs in $O(3^{n})$.
\end{thm}

\begin{proof}
SEFAR runs in parallel several processes, each with a different set of sizes, and terminates when all subspaces are fully evaluated. To evaluate the coalitions, each process uses dynamic programming. The worst-case run time of dynamic programming on this problem is $\mathcal{O}(3^{n})$~\cite{yeh1986dynamic}. Hence, the time complexity of SEFAR is $\mathcal{O}(3^{n})$.
\end{proof}

\begin{thm}
For all CSG problem sizes $n$, the preprocessor of SEFAR computes one set of sizes in $\mathcal{O}(2^{n} \times n^{2} \times \frac{e^{\pi\sqrt{\frac{2n}{3}}}}{n})$.
\end{thm}

\begin{proof}
For each problem size $n$, the preprocessor evaluates all possible sets of coalition sizes. For each set of coalition sizes, it constructs the integer partition graph by dividing only the integers that belong to that set and tests whether it is beneficial to consider that set or not. The total number of sets that the preprocessor evaluates is $2^{n-2}-1$. We denote by $\mathcal{I}(s)$ the number of integer splits performed for a set $s$ and by $p(n,s)$, the number of subspaces generated by the set $s$.

Total operations $\mathcal{T}(n)$ performed by the preprocessor is as follows:\\
$\mathcal{T}(n) = \sum_{s=1}^{2^{n-2}-1} \mathcal{I}(s)$\\
$\mathcal{T}(n) = \sum_{s=1}^{2^{n-2}-1} \sum_{\mathcal{N}} \mathcal{I}(\mathcal{N},s)$, where $\mathcal{I}(\mathcal{N},s)$ is the number of integer splits performed on the node $\mathcal{N}$.\par
Notice that the number of splits into two for a certain integer $i$ is $\left\lfloor\frac{i}{2}\right\rfloor$, the highest integer to split is $n$ and the highest possible number of integers in a single node is $n$, which is the number of integers of the node that represents the singleton coalition structure. Thus, $\mathcal{I}(\mathcal{N},s) \leq n \times \left\lfloor\frac{n}{2}\right\rfloor$\\
$\mathcal{T}(n) \leq \sum_{s=1}^{2^{n-2}-1} p(n,s) \times n \times \left\lfloor\frac{n}{2}\right\rfloor$. However, the growth rate of the number of nodes in the integer partition graph, which is the same as the growth rate of the integer partition of $n$, is known to be $\mathcal{O}(\frac{e^{\pi\sqrt{\frac{2n}{3}}}}{n})$~\cite{herbert2000lectures}. Hence:\\
$\mathcal{T}(n) \leq \sum_{s=1}^{2^{n-2}-1} \mathcal{O}(\frac{e^{\pi\sqrt{\frac{2n}{3}}}}{n}) \times n \times \left\lfloor\frac{n}{2}\right\rfloor$\\
$\mathcal{T}(n) \leq 2^{n-2}-1 \times \mathcal{O}(\frac{e^{\pi\sqrt{\frac{2n}{3}}}}{n}) \times n \times \left\lfloor\frac{n}{2}\right\rfloor$\\
As a result, the time complexity of the preprocessor to compute one set of sizes is:\\
$\mathcal{O}(n^{2} \times 2^{n} \times \frac{e^{\pi\sqrt{\frac{2n}{3}}}}{n})$

The preprocessor is run only once and the resulting sets are reused for each CSG problem-solving.
\end{proof}

\section{Empirical Evaluation}

\begin{figure*}[t]
\begin{center}
\small
\resizebox{0.31\textwidth}{4.8cm}{%
     \begin{subfigure}{0.28\textwidth}
         \centering
         \begin{tikzpicture}
	\begin{loglogaxis}[	grid= major ,
            title=Normal ,
			width=0.8\textwidth ,
            xlabel = {Number of agents} ,
			ylabel = {Time (in seconds)} ,
            width=5cm,height=5cm,
            xtick={15,16,17,18,19,20,21,22},
            xticklabels={15,16,17,18,19,20,21,22,23,24,25,26},
            ymode=log,
            ymin=0,
            label style={font=\large},
			title style={font=\Large},
			tick label style={font=\footnotesize},
            legend entries={SEFAR, ODP-IP, BOSS},
			legend style={at={(0.8,1.6)},anchor=north}]
			\addplot+[ultra thin,mark size=1pt] coordinates {(15,0.033) (16,0.066) (17, 0.128) (18,0.209) (19,0.474) (20,1.02) (21,3.0) (22,6.1) };//
			\addplot+[ultra thin,mark size=1pt] coordinates {(15,0.055) (16,0.086) (17, 0.122) (18,0.221) (19,0.498) (20,1.2) (21,3.0) (22,5.2) };
            \addplot+[ultra thin,mark size=1pt,vert] coordinates {(15,0.063) (16,0.091) (17, 0.143) (18,0.205) (19,0.523) (20,1.2) (21,3.2) (22,4.9) };
            
	\end{loglogaxis}
    \end{tikzpicture}\\
         \label{fig:three sin x}
     \end{subfigure}
     }
\resizebox{0.31\textwidth}{4.8cm}{%
     \begin{subfigure}{0.28\textwidth}
         \centering
         \begin{tikzpicture}
	\begin{loglogaxis}[	grid= major ,
            title=Agent-based Normal,
			width=0.6\textwidth ,
            xlabel = {Number of agents} ,
			ylabel = {Time (in seconds)} ,
            width=5cm,height=5cm,
            xtick={15,16,17,18,19,20,21,22},
            xticklabels={15,16,17,18,19,20,21,22,23,24,25,26},
            ymode=log,
            label style={font=\large},
			title style={font=\Large},
			tick label style={font=\footnotesize},
            legend entries={.,.,.},
			legend style={at={(1.75,1.5)},anchor=north}]
			\addplot+[ultra thin,mark size=1pt] coordinates {(15,0.034) (16,0.063) (17, 0.144) (18,0.232) (19,0.423) (20,1.09) (21,3.1) (22,6.9) };//
			\addplot+[ultra thin,mark size=1pt] coordinates {(15,0.065) (16,0.101) (17, 0.179) (18,0.303) (19,0.749) (20,1.61) (21,4.0) (22,17.042) };
			\addplot+[ultra thin,mark size=1pt,vert] coordinates {(15,0.053) (16,0.091) (17, 0.154) (18,0.262) (19,0.472) (20,1.25) (21,2.7) (22,9.2) };
	\end{loglogaxis}
    \end{tikzpicture}\\
         \label{fig:y equals x}
     \end{subfigure}
     }
     \resizebox{0.31\textwidth}{4.8cm}{%
     \begin{subfigure}{0.28\textwidth}
         \centering
         \begin{tikzpicture}
	\begin{loglogaxis}[	grid= major ,
            title=Agent-based Uniform ,
			width=0.8\textwidth ,
            xlabel = {Number of agents} ,
			ylabel = {Time (in seconds)} ,
            width=5cm,height=5cm,
            xtick={15,16,17,18,19,20,21,22},
            xticklabels={15,16,17,18,19,20,21,22,23,24,25,26},
            ymode=log,
            label style={font=\large},
			title style={font=\Large},
			tick label style={font=\footnotesize},
            legend entries={SEFAR, ODP-IP, BOSS},
			legend style={at={(0.2,1.6)},anchor=north}]
			\addplot+[ultra thin,mark size=1pt] coordinates {(15,0.035) (16,0.067) (17, 0.142) (18,0.228) (19,0.513) (20,1.06) (21,3.4) (22,6.7) }; //
			\addplot+[ultra thin,mark size=1pt] coordinates {(15,0.059) (16,0.097) (17, 0.190) (18,0.320) (19,0.732) (20,1.87) (21,4.1) (22,17.1) };
			\addplot+[ultra thin,mark size=1pt,vert] coordinates {(15,0.055) (16,0.089) (17, 0.151) (18,0.304) (19,0.498) (20,1.37) (21,2.9) (22,9.9) };
	\end{loglogaxis}
    \end{tikzpicture}\\
         \label{fig:three sin x}
     \end{subfigure}
     }
     \hfill
     \resizebox{0.31\textwidth}{4.8cm}{%
     \begin{subfigure}{0.28\textwidth}
         \centering
         \begin{tikzpicture}
	\begin{loglogaxis}[	grid= major ,
            title=Modified Uniform ,
			width=0.8\textwidth ,
            xlabel = {Number of agents} ,
			ylabel = {Time (in seconds)} ,
            width=5cm,height=5cm,
            xtick={15,16,17,18,19,20,21,22},
            xticklabels={15,16,17,18,19,20,21,22},
            ymode=log,
            label style={font=\large},
			title style={font=\Large},
			tick label style={font=\footnotesize},
            legend entries={SEFAR, ODP-IP, BOSS},
			legend style={at={(0.8,1.6)},anchor=north}]
			\addplot+[ultra thin,mark size=1pt] coordinates {(15,0.034) (16,0.045) (17,0.091)  (18,0.147) (19,0.421) (20,0.64) (21,1.7) (22,3.8) };//
			\addplot+[ultra thin,mark size=1pt] coordinates {(15,0.049) (16,0.079) (17,0.137)  (18,0.182) (19,0.297) (20,0.72) (21,3.3) (22,4.6) };
			\addplot+[ultra thin,mark size=1pt,vert] coordinates {(15,0.053) (16,0.072) (17,0.131)  (18,0.192) (19,0.389) (20,1.2) (21,3.7) (22,4.3) };
	\end{loglogaxis}
    \end{tikzpicture}\\
         \label{fig:three sin x}
     \end{subfigure}
     }
     \resizebox{0.31\textwidth}{4.8cm}{%
     \begin{subfigure}{0.28\textwidth}
         \centering
         \begin{tikzpicture}
	\begin{loglogaxis}[	grid= major ,
            title=Modified Normal ,
			width=0.8\textwidth ,
            xlabel = {Number of agents} ,
			ylabel = {Time (in seconds)} ,
            width=5cm,height=5cm,
            xtick={15,16,17,18,19,20,21,22},
            xticklabels={15,16,17,18,19,20,21,22,23,24,25,26},
            ymode=log,
            label style={font=\large},
			title style={font=\Large},
			tick label style={font=\footnotesize},
            legend entries={.,.,.},
			legend style={at={(1.75,1.5)},anchor=north}]
			\addplot+[ultra thin,mark size=1pt] coordinates {(15,0.028) (16,0.055) (17, 0.109) (18,0.165) (19,0.301) (20,0.59) (21,1.5) (22,2.7) };//? 
			\addplot+[ultra thin,mark size=1pt] coordinates {(15,0.049) (16,0.067) (17, 0.099) (18,0.100) (19,0.178) (20,0.35) (21,1.09) (22,1.254) };
			\addplot+[ultra thin,mark size=1pt,vert] coordinates {(15,0.055) (16,0.081) (17, 0.115) (18,0.150) (19,0.284) (20,0.95) (21,1.87) (22,2.154) };
	\end{loglogaxis}
    \end{tikzpicture}\\
         \label{fig:three sin x}
     \end{subfigure}
     }
     \resizebox{0.31\textwidth}{4.8cm}{%
     \begin{subfigure}{0.28\textwidth}
         \centering
         \begin{tikzpicture}
	\begin{loglogaxis}[	grid= major ,
            title=Pascal ,
			width=0.8\textwidth ,
            xlabel = {Number of agents} ,
			ylabel = {Time (in seconds)} ,
            width=5cm,height=5cm,
            xtick={15,16,17,18,19,20,21,22},
            xticklabels={15,16,17,18,19,20,21,22,23,24,25,26},
            ymode=log,
            label style={font=\large},
			title style={font=\Large},
			tick label style={font=\footnotesize},
            legend entries={SEFAR, ODP-IP, BOSS},
			legend style={at={(0.2,1.6)},anchor=north}]
			\addplot+[ultra thin,mark size=1pt] coordinates {(15,0.015) (16,0.037) (17, 0.049) (18,0.091) (19,0.072) (20,0.112) (21,0.179) (22,0.286) }; 
			\addplot+[ultra thin,mark size=1pt] coordinates {(15,0.029) (16,0.049) (17, 0.061) (18,0.094) (19,0.089) (20,0.111) (21,0.194) (22,0.412) };
			\addplot+[ultra thin,mark size=1pt,vert] coordinates {(15,0.030) (16,0.043) (17, 0.064) (18,0.102) (19,0.101) (20,0.136) (21,0.265) (22,0.414) };
	\end{loglogaxis}
    \end{tikzpicture}\\
         \label{fig:three sin x}
     \end{subfigure}
}
\hfill
\resizebox{0.31\textwidth}{4.8cm}{%
     \begin{subfigure}{0.28\textwidth}
         \centering
         \begin{tikzpicture}
	\begin{loglogaxis}[	grid= major ,
            title=Weibull ,
			width=0.8\textwidth ,
            xlabel = {Number of agents} ,
			ylabel = {Time (in seconds)} ,
            width=5cm,height=5cm,
            xtick={15,16,17,18,19,20,21,22},
            xticklabels={15,16,17,18,19,20,21,22,23,24,25,26},
            ymode=log,
            label style={font=\large},
			title style={font=\Large},
			tick label style={font=\footnotesize},
            legend entries={SEFAR, ODP-IP, BOSS},
			legend style={at={(0.8,1.6)},anchor=north}]
			\addplot+[ultra thin,mark size=1pt] coordinates {(15,0.029) (16,0.061) (17, 0.092) (18,0.186) (19,0.455) (20,0.71) (21,1.8) (22,3.4) };// 
			\addplot+[ultra thin,mark size=1pt] coordinates {(15,0.065) (16,0.091) (17, 0.177) (18,0.224) (19,0.460) (20,0.72) (21,3.3) (22,2.7) };
			\addplot+[ultra thin,mark size=1pt,vert] coordinates {(15,0.063) (16,0.068) (17, 0.172) (18,0.219) (19,0.425) (20,1.2) (21,3.7) (22,2.9) };
	\end{loglogaxis}
    \end{tikzpicture}\\
         \label{fig:three sin x}
     \end{subfigure}
     }
     \resizebox{0.31\textwidth}{4.8cm}{%
     \begin{subfigure}{0.28\textwidth}
         \centering
         \begin{tikzpicture}
	\begin{loglogaxis}[	grid= major ,
            title=Gamma ,
			width=0.8\textwidth ,
            xlabel = {Number of agents} ,
			ylabel = {Time (in seconds)} ,
            width=5cm,height=5cm,
            xtick={15,16,17,18,19,20,21,22},
            xticklabels={15,16,17,18,19,20,21,22,23,24,25,26},
            ymode=log,
            label style={font=\large},
			title style={font=\Large},
			tick label style={font=\footnotesize},
            legend entries={.,.,.,.},
			legend style={at={(1.75,1.5)},anchor=north}]
			\addplot+[ultra thin,mark size=1pt] coordinates {(15,0.028) (16,0.063) (17, 0.124) (18,0.243) (19,0.438) (20,0.97) (21,3.0) (22,4.9) }; //
			\addplot+[ultra thin,mark size=1pt] coordinates {(15,0.046) (16,0.086) (17, 0.186) (18,0.226) (19,0.508) (20,0.84) (21,1.6) (22,3.5) };
			\addplot+[ultra thin,mark size=1pt,vert] coordinates {(15,0.048) (16,0.097) (17, 0.195) (18,0.217) (19,0.485) (20,1.08) (21,1.7) (22,3.8) };
	\end{loglogaxis}
    \end{tikzpicture}\\
         \label{fig:three sin x}
     \end{subfigure}
     }
     \resizebox{0.31\textwidth}{4.8cm}{%
     \begin{subfigure}{0.28\textwidth}
         \centering
         \begin{tikzpicture}
	\begin{loglogaxis}[	grid= major ,
            title=Exponential ,
			width=0.8\textwidth ,
            xlabel = {Number of agents} ,
			ylabel = {Time (in seconds)} ,
            width=5cm,height=5cm,
            xtick={15,16,17,18,19,20,21,22},
            xticklabels={15,16,17,18,19,20,21,22,23,24,25,26},
            ymode=log,
            label style={font=\large},
			title style={font=\Large},
			tick label style={font=\footnotesize},
            legend entries={SEFAR, ODP-IP, BOSS},
			legend style={at={(0.2,1.6)},anchor=north}]
			\addplot+[ultra thin,mark size=1pt] coordinates {(15,0.029) (16, 0.039) (17, 0.093) (18, 0.164) (19, 0.343) (20,0.95) (21,3.1) (22,5.7) }; 
			\addplot+[ultra thin,mark size=1pt] coordinates {(15,0.053) (16, 0.084) (17, 0.203) (18, 0.201) (19, 0.545) (20,0.75) (21,1.9) (22,4.4) };
			\addplot+[ultra thin,mark size=1pt,vert] coordinates {(15,0.058) (16, 0.078) (17, 0.199) (18, 0.197) (19, 0.512) (20,0.7) (21,2.1) (22,4.3) };
	\end{loglogaxis}
    \end{tikzpicture}\\
         \label{fig:three sin x}
     \end{subfigure}
}

             \caption{Time performance in seconds of SEFAR, ODP-IP and BOSS for a number of agents between 15 and 22. A lower run time is better.
             }
        \label{sefarResults}
\normalsize
\end{center}
\end{figure*}
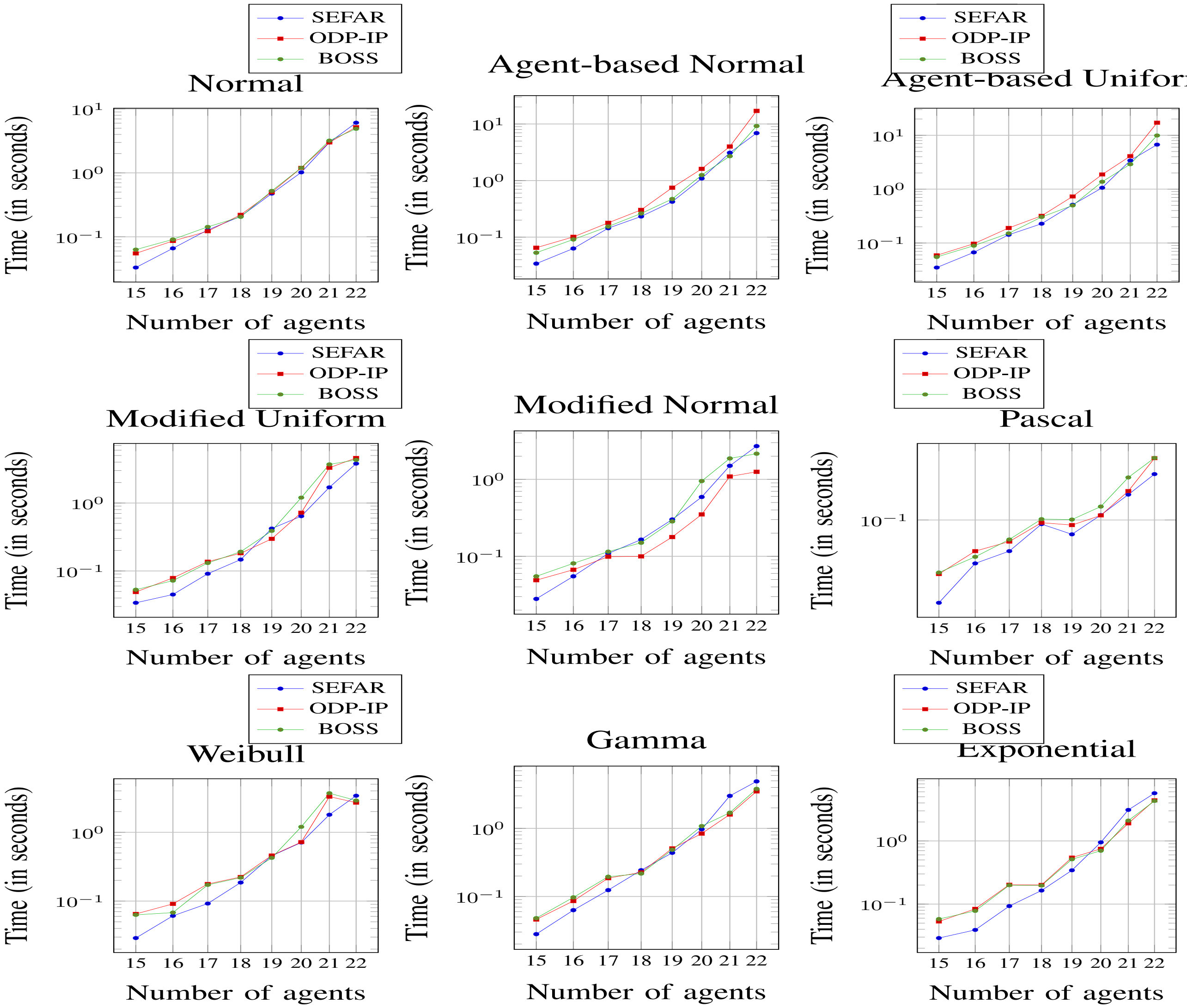

Our main goals with the empirical evaluation are to investigate: 1) how different input-sizes, value distributions and time constraints affect the algorithm performance; 2) the algorithm’s performance relative to other dynamic programming algorithms; 3) the algorithm’s performance relative to other state-of-the-art methods. To do this, we use the following common coalition value distributions to generate problem instances: Agent-based Uniform, Modified Normal~\cite{rahwan2012hybrid}, Agent-based Normal, Exponential, Gamma~\cite{michalak2016hybrid}, Normal~\cite{rahwan2007near}, Modified Uniform~\cite{service2010approximate}, Pascal, and Weibull~\cite{changder2020odss}. 
The result for each value distribution was produced by computing the average of the resulting values from 50 generated problem instances per value distribution. All algorithms were implemented in Java and we used the codes provided by the authors of the tested algorithms. The parameter optimization and the algorithms were run on an Intel Xeon 2.30GHz E5-2650 CPU with 256GB of RAM.

To see how the preprocessor affects the performance of our algorithm, we benchmark against the dynamic programming algorithms IDP~\cite{rahwan2008improved} and ODP~\cite{michalak2016hybrid}. In these experiments, we show the worst-case time results of SEFAR and the run times of IDP and ODP. To obtain the worst-case run time for SEFAR, we omit the subspace pruning using the upper bound (line 10 in Algorithm \ref{gradp}, which is described in Section 3.2), meaning that SEFAR finishes and returns the optimal solution when all of its processes finish. We also show the time gain achieved by SEFAR compared to IDP. Recall that the performance of IDP and ODP does not depend on the value distribution, but only on the number of agents.

Table \ref{tabresult} clearly shows that SEFAR provides optimal solutions faster than IDP and ODP. For example, after 1 second, at worst case for 20 agents, SEFAR returns optimal solutions roughly $73\%$ faster than IDP and $55\%$ faster than ODP. This behavior is consistent with the other numbers of agents. This shows that the preprocessor, while focusing on the subspaces, enhances the effectiveness of the algorithm.


\begin{table}[h!]
\begin{center}
\renewcommand{\arraystretch}{1.4}
\begin{tabular}{|c|c|c|c|c|}
\hline
\textbf{Number of Agents} & \multicolumn{3}{c|}{\textbf{Execution Time}} &  \textbf{Time} \\
\cline{2-4}
 &  \textbf{SEFAR} & \textbf{IDP} & \textbf{ODP} & \textbf{Gain} \\
\hline
15 & $0.011$ & $0.033$ & $0.026$ & $67\%$ \\
\hline
16 & $0.022$ & $0.072$ & $0.058$ & $70\%$ \\
\hline
17 & $0.040$ & $0.167$ & $0.128$ & $77\%$ \\
\hline
18 & $0.121$ & $0.520$ & $0.398$ & $77\%$ \\
\hline
19 & $0.470$ & $1.416$ & $1.076$ & $67\%$ \\
\hline
20 & $1.0$ & $3.7$ & $2.2$ & $73\%$ \\
\hline
21 & $4.8$ & $15.9$ & $10.3$ & $70\%$ \\
\hline
22 & $7.9$ & $24.7$ & $19.8$ & $69\%$ \\
\hline
\end{tabular}
\end{center}
\caption{Worst-case run time in seconds of SEFAR compared to the run time of IDP, and ODP. The `Time  Gain' column shows the time gain achieved by SEFAR compared to IDP.}\label{tabresult}
\end{table}

To see how our algorithm, using the upper-bound subspace pruning, behave against the state-of-the-art algorithms ODP-IP~\cite{michalak2016hybrid} and BOSS~\cite{Changder_Aknine_Ramchurn_Dutta_2021}, we plot the results from different benchmarks in Fig. \ref{sefarResults}. Note that the algorithms behave differently depending on the value distributions. Overall, SEFAR performs better than ODP-IP and BOSS in some value distributions and on par on other value distributions. Moreover, our results show that the time required for ODP-IP and BOSS for optimally solving a problem exceeds in many cases the worst-case run time of SEFAR shown in Table \ref{tabresult}.
A notable exception is the Modified Normal distribution, for which the state-of-the-art algorithm ODP-IP outperforms the proposed algorithm SEFAR, and the Normal distribution, for which there is almost a tie. An important observation is that there is no distribution for which SEFAR performs much worse than the state-of-the-art algorithms—--indicating that our algorithm is potentially suitable for many different problem instances, including those of easily solvable distribution instances, such as Pascal and Modified Uniform, and those of difficult distribution instances, such as Agent-based Normal and Agent-based Uniform.

\section{Conclusion}

In this paper, we presented the first algorithm based on parameter configuration for optimally solving the coalition structure generation problem. In more detail, we developed a preprocessor that selects the best sets of parameters for dynamic programming, and then introduced a new algorithm that searches the solution space using dynamic programming with a branch-and-bound technique both focused on solution subspaces. This algorithm can be used for solving many combinatorial problems where dynamic programming is used. We ran experiments on a variety of different value distributions and our results demonstrate that our algorithm can perform on par or even better than existing well established state-of-the-art algorithms in solving the coalition structure generation problem.



\end{document}